\documentclass[runningheads]{llncs}
\newlength{\origtabcolsep}
\usepackage{amsmath,amssymb, amsfonts, stmaryrd}
\usepackage{hyperref}
\usepackage{multirow} 
\usepackage{wrapfig}
\usepackage{newclude}

\usepackage{bm}
\usepackage{xcolor}   
\usepackage{amssymb}  

\usepackage{makecell} 
\usepackage{booktabs}
\usepackage{array}
\usepackage{marvosym}
\DeclareRobustCommand{\corr}{\textsuperscript{\scriptsize\Letter}}
\usepackage{xcolor}
\usepackage{pifont}
\usepackage{graphicx}
\usepackage{colortbl}

\begin{document}
\title{PRA-PoE: Robust Multimodal \\ Alzheimer's Disease Classification under \\Arbitrary Modality Missingness}

\titlerunning{PRA-PoE: Robust Alzheimer's Diagnosis with Arbitrary Missing Modalities}
%
%

\author{Guangqian Yang\inst{1} 
\and
Ye Du\inst{1} 
\and 
Wenlong Hou\inst{1} 
\and 
Qian Niu\inst{2}\corr 
\and 
Shujun Wang\inst{1,3}\corr 
\and 
for the Alzheimer's Disease Neuroimaging Initiative}
\authorrunning{G. Yang et al.}
%
\institute{Department of Biomedical Engineering, The Hong Kong Polytechnic University, Hong Kong
SAR, China
\and Department of Technology Management for Innovation, The University of Tokyo, Japan
\and Department of Data Science and Artificial Intelligence, The Hong Kong Polytechnic University, Hong Kong SAR, China \\
\email{qian.niu@weblab.t.u-tokyo.ac.jp, shu-jun.wang@polyu.edu.hk}\\
\Letter~Corresponding authors.}

\maketitle              
\begin{abstract}
Missing modalities are prevalent in real-world Alzheimer’s disease (AD) assessment and pose a significant challenge to multimodal learning, particularly when the distribution of observed modality subsets differs between training and deployment. Such missingness pattern mismatch induces a conditional representation shift across modality subsets. Existing approaches that rely on implicit imputation or modality synthesis often fail to explicitly model modality availability and uncertainty, leading to overconfident dependence on synthesized features, reduced robustness, and miscalibrated uncertainty estimates.
To address these limitations, we propose PRA-PoE, an incomplete multimodal learning framework that is equipped with Prototype-anchored Representation Alignment (PRA) and an Uncertainty-aware Product of Experts (UA-PoE) fusion mechanism. First, PRA uses learnable global prototypes and availability-conditioned tokens to encode modality availability, distinguish observed from missing modalities, re-synthesize features for missing modalities, and adaptively refine observed representations to align latent spaces across modality subsets, with the goal of reducing representation shift under varying missingness patterns. Second, UA-PoE models each modality as a Gaussian expert and performs closed-form Product of Experts fusion, where experts with higher uncertainty are automatically down-weighted via lower precision, improving uncertainty reliability.
We evaluate PRA-PoE under a clinically realistic protocol by training with naturally missing data and testing on all non-empty modality combinations. PRA-PoE consistently outperforms the state-of-the-art across datasets, achieving a \textbf{5.4\%} relative improvement in average accuracy on ADNI and a \textbf{10.9\%} relative gain in average F1 on OASIS-3 over the strongest baseline across all non-empty modality subsets. Code is available at \url{https://github.com/SDH-Lab/PRA-PoE}.

\end{abstract}

\section{Introduction}
\begin{figure}[h]
\centerline{\includegraphics[width=0.9\columnwidth]{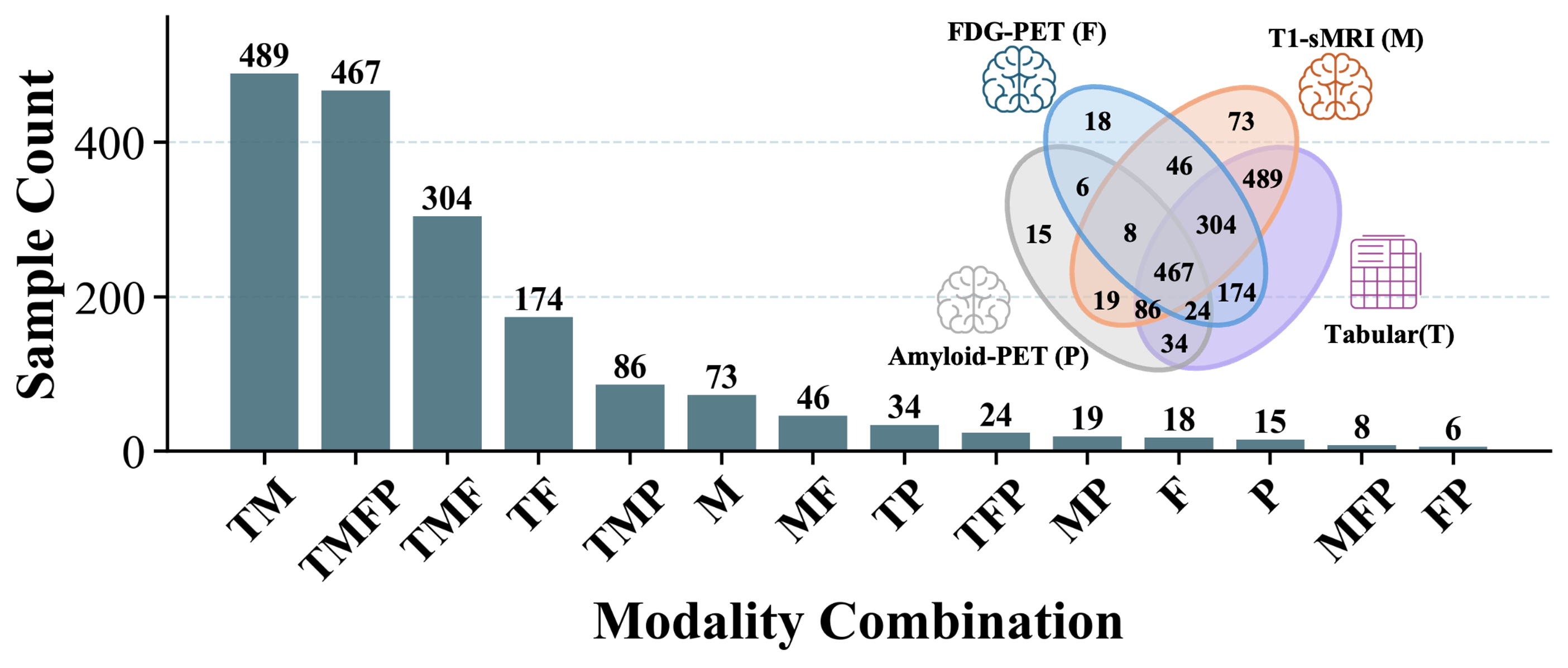}}
\caption{Modality missing patterns and sample counts for each combination in the ADNI dataset~\cite{mueller2005alzheimer}. 
} 
\label{fig:intro}
\end{figure}

Accurate classification of Alzheimer’s disease (AD) relies on multimodal evidence \cite{lei2024alzheimer, hou2025adagent, yang2025adfound}. Structural Magnetic Resonance Imaging (sMRI), Positron Emission Tomography (PET) (e.g., FDG-PET and Amyloid-PET), and clinical assessments provide complementary cues on anatomy, metabolism/pathology~\cite{minoshima202218f, xie2023image}, and symptoms, making multimodal learning a key paradigm for AD staging. However, in real-world clinical cohorts, multimodal acquisition is routinely incomplete due to cost, scanner availability, patient burden, contraindications, and protocol heterogeneity \cite{kwak2025cross, odusami2024alzheimer}, as illustrated in Fig.~\ref{fig:intro}. This missingness is largely protocol-driven. Tabular data (e.g., demographic information and cognitive assessments) are routinely available, while MRI/PET can be absent due to contraindications, radiation concerns, and cost~\cite{kwak2025cross, odusami2024alzheimer}, yielding non-random and highly imbalanced modality subsets. Consequently, modality availability varies across subjects, and models must remain accurate and reliable under heterogeneous missingness.

To learn under heterogeneous modality availability, prior work on incomplete multimodal learning mainly follows two lines: (i) missingness-robust deterministic fusion, e.g., modality dropout~\cite{feng2024unified, zhang2022mmformer}, mixture-of-fusion strategies~\cite{yun2025flex, han2025fusemoe,yun2025generate}, and knowledge distillation~\cite{kwak2025cross}; and (ii) imputation or cross-modal translation that predicts missing modalities or their features from observed ones~\cite{zhang2022m3care,meng2024multi, chen2023disentangle, pan2021disease, pan2018synthesizing}. Yet clinical AD cohorts exhibit a long-tailed distribution over observed modality subsets (Fig.~\ref{fig:intro}) due to hierarchical acquisition protocols, creating a missingness pattern mismatch between the dominant subsets seen during training and the rare subsets encountered at deployment.
This mismatch raises two challenges. First, models tend to co-adapt to frequent modality subsets, so the representation of a modality can vary with its surrounding context (i.e., which other modalities are present), leading to missingness-induced representation shift and degraded generalization on rare subsets \cite{zhao2025mce, dai2025unbiased}. Second, although imputation-based approaches can yield uncertain feature estimates, the previous methods~\cite{zhang2022m3care, chen2023disentangle} often treat synthesized features as if they were fully observed, lacking an explicit mechanism to down-weight unreliable estimates. Such blind fusion can induce negative transfer, where noisy imputations corrupt multimodal aggregation and reduce performance even when strong observed modalities are available.

To address these challenges, we propose Prototype-anchored Representation Alignment with Product-of-Experts fusion (PRA-PoE), a unified framework for robust multimodal learning under missing data. PRA introduces learnable global prototypes together with availability-conditioned tokens to (i) align modality representations into a subset-consistent space, reducing context-dependent representation shift, and (ii) synthesize missing modality features from the observed ones at the feature level. Building on these representations, UA-PoE treats each modality as a Gaussian expert and performs adaptive Product-of-Experts fusion, where experts with higher uncertainty contribute less via lower precision. This explicitly prevents overconfident reliance on synthesized features and mitigates negative transfer under severe missingness. We evaluate PRA-PoE under a clinically realistic protocol that trains with naturally missing modalities and tests on all non-empty modality subsets, including rare long-tailed patterns. Extensive experiments on two large-scale benchmarks demonstrate the effectiveness of PRA-PoE, which surpasses the strongest baseline by relative improvements of 5.4\% in average accuracy on ADNI and 10.9\% in average F1 on OASIS-3 under arbitrary missing-modality scenarios.

\section{Method}

\subsection{Overview and Problem Formulation}

\textbf{Overview.}
We study AD stage classification with incomplete multimodal inputs. Training follows the naturally missing modalities, while testing must work with any non-empty observed subset. This subset-dependent input creates a representation shift and can cause negative transfer when missing modalities are completed with overconfident features. We propose \textbf{PRA-PoE}: modality encoders map inputs to a shared latent space, PRA performs subset-conditioned prototype-anchored alignment and completion, and UA-PoE fuses modalities by Gaussian product of experts with uncertainty-aware down-weighting, followed by Monte Carlo sampling for prediction. 

\noindent\textbf{Problem formulation.}
Let $\mathcal{D}=\{(X_i,y_i)\}_{i=1}^{N}$, where $y_i\in\{\mathrm{CN},\mathrm{MCI},\mathrm{AD}\}$ and
$X_i=\{x_i^{m}\}_{m=1}^{M}$ contains $M=4$ modalities: T1-sMRI, FDG-PET, Amyloid-PET, and tabular clinical or demographic variables.
We use an observation mask $\delta_{i,m}\in\{0,1\}$ and define the observed set
$\mathcal{O}_i=\{m \mid \delta_{i,m}=1\}$.
Tabular variables are usually present but can be missing, while imaging modalities are often absent, which yields imbalanced modality subsets.

\subsection{Unified Feature Encoding}

We employ modality-specific encoders \(E_m(\cdot)\) to map the heterogeneous inputs into a shared $D$-dimensional space, defined as $\mathbf{h}_{i,m}=E_m(x_i^m)\in\mathbb{R}^{D}$.
For 3D neuroimaging modalities (T1-sMRI, FDG-PET, Amyloid-PET), \(E_m(\cdot)\) is a 3D volume encoder; for tabular data, \(E_m(\cdot)\) is an MLP on standardized variables. 
When modality $m$ is missing, we set $\mathbf{h}_{i,m}=\mathbf{0}$ to maintain tensor-shape consistency. These zero placeholders do not contribute to cross-modal aggregation, as they are excluded from attention via an availability mask in PRA (Sec.~\ref{sec:pra}).

\begin{figure}[!t]
\centerline{\includegraphics[width=1\columnwidth]{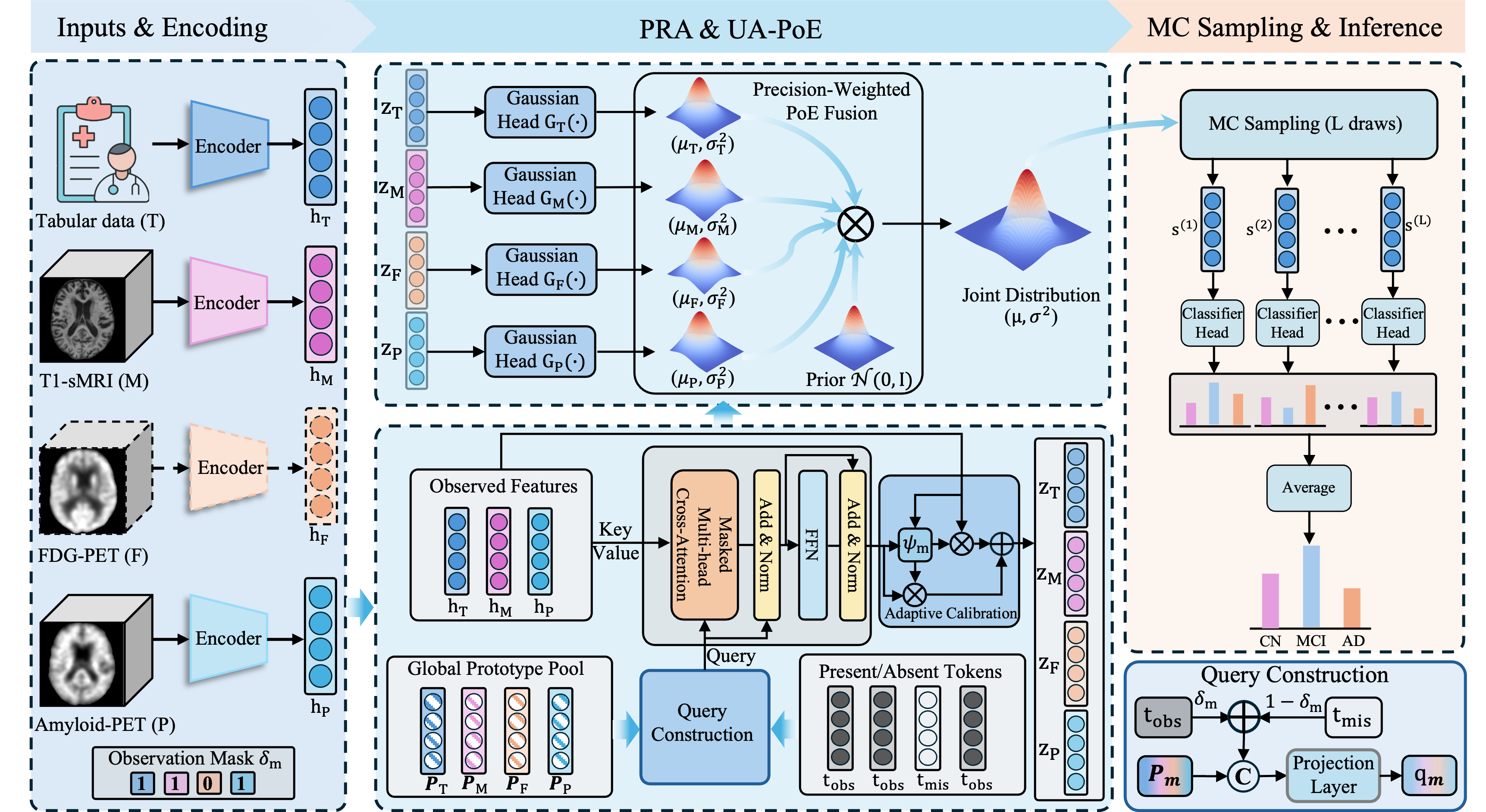}}
\caption{Overall Framework of PRA-PoE. 
} 

\label{fig:framework}
\end{figure}
\subsection{Prototype-Anchored Representation Alignment (PRA)}
\label{sec:pra}
Direct fusion of incomplete multimodal features leads to \emph{missingness-induced representation shift}, where a modality’s embedding becomes dependent on the specific subset of co-observed modalities, resulting in context-dependent drift and unstable alignment.
To mitigate this issue, we propose \textbf{PRA}, which enforces subset-invariant semantics prior to fusion. PRA maps each modality into a shared semantic space by anchoring representations to learnable global prototypes and refining them via availability-conditioned cross-modal attention. Applied to all modalities, including observed ones, \textbf{PRA} explicitly corrects subset-induced drift by constraining representations toward global anchors while leveraging the available cross-modal context.


\noindent\textbf{Prototype-anchored Query with Availability Embedding.} For each modality \(m\), we use a learnable global prototype \(\mathbf{P}_m\in\mathbb{R}^{D}\) and two learnable availability-conditioned tokens \(\mathbf{t}_{\mathrm{obs}},\mathbf{t}_{\mathrm{mis}}\in\mathbb{R}^{D}\) to construct query for each modality \(m\), as follows:
\begin{equation}
\mathbf{a}_{i,m}=\delta_{i,m}\mathbf{t}_{\mathrm{obs}}+(1-\delta_{i,m})\mathbf{t}_{\mathrm{mis}},
\qquad
\mathbf{q}_{i,m}=\phi_m\!\left([\mathbf{P}_m \parallel \mathbf{a}_{i,m}]\right)\in\mathbb{R}^{D},
\end{equation}
where $\phi_m:\mathbb{R}^{2D}\rightarrow\mathbb{R}^{D}$ is a linear layer and $[\cdot\parallel\cdot]$ denotes concatenation.

\noindent\textbf{Masked Multi-head Cross-Attention.}
For sample $i$ and target modality $m$, we stack the features of the remaining $M{-}1$ modalities into a context matrix
$\mathbf{C}_{i,m}=[\mathbf{h}_{i,m_1};\dots;\mathbf{h}_{i,m_{M-1}}]\in\mathbb{R}^{(M-1)\times D}$,
where $\{m_1,\dots,m_{M-1}\}=\{1,\dots,M\}\setminus\{m\}$.
The prototype-anchored query $\mathbf{q}_{i,m}$ serves as the query, and $\mathbf{C}_{i,m}$ serves as both keys and values.
To prevent attention to missing modalities, we derive an additive bias $\mathbf{B}_{i,m}\in\mathbb{R}^{1\times(M-1)}$ directly from the observation mask:
\begin{equation}
(\mathbf{B}_{i,m})_j =
\begin{cases}
0, & \text{if } \delta_{i,m_j}=1,\\
-\infty, & \text{otherwise},
\end{cases}
\end{equation}
and compute the refined representation via multi-head cross-attention:
\begin{equation}
\hat{\mathbf{h}}_{i,m}
= \mathrm{softmax}\!\left(\frac{\mathbf{q}_{i,m}\,\mathbf{C}_{i,m}^{\top}}{\sqrt{d_k}}+\mathbf{B}_{i,m}\right)\mathbf{C}_{i,m},
\end{equation}
where $d_k$ is the per-head dimension.
This formulation maintains a fixed $(M{-}1)$-token context shape regardless of the missing pattern, while ensuring that only observed modalities contribute to cross-modal refinement.

\noindent\textbf{Adaptive Calibration and PRA Output.}
For observed modalities, we adaptively fuse the original feature \(\mathbf{h}_{i,m}\) with the aligned feature \(\hat{\mathbf{h}}_{i,m}\):
\begin{equation}
\alpha_{i,m}=\sigma\!\left(\psi_m([\mathbf{h}_{i,m}\parallel \hat{\mathbf{h}}_{i,m}])\right),
\qquad
\mathbf{f}_{i,m}=\alpha_{i,m}\mathbf{h}_{i,m}+(1-\alpha_{i,m})\hat{\mathbf{h}}_{i,m},
\end{equation}
where \(\psi_m\) is a linear layer. The final aligned representation is
\begin{equation}
\mathbf{z}_{i,m}=\mathrm{LN}\!\left(\delta_{i,m}\mathbf{f}_{i,m}+(1-\delta_{i,m})\hat{\mathbf{h}}_{i,m}\right).
\label{eq:zm_def}
\end{equation}

Thus, missing modalities never serve as cross-modal information sources; their \(\mathbf{z}_{i,m}\) are inferred from \(\mathcal{O}_i\) via \(\hat{\mathbf{h}}_{i,m}\), while observed modalities are calibrated to reduce subset-induced drift.

\subsection{Uncertainty-Aware Product-of-Experts (UA-PoE)}
\label{sec:uapoe}

Given the PRA-aligned features $\{\mathbf{z}_{i,m}\}_{m=1}^{M}$, we
formulate multimodal fusion as Gaussian PoE in a shared latent
space~\cite{shi2019variational}. The standard strategy for
handling missing modalities removes absent experts from the product, fusing
only over $\mathcal{O}_i$. In contrast, we retain all $M$ experts, including
those whose features are completed by PRA, and let each expert independently
predict its own mean and variance. Because PRA-completed features carry
higher epistemic uncertainty, their corresponding experts learn larger
variances and thus contribute less precision to the fused posterior,
achieving implicit uncertainty-aware down-weighting within the closed-form
PoE framework.

Concretely, a lightweight head $G_m(\cdot)$ maps each $\mathbf{z}_{i,m}$ to
a diagonal Gaussian expert
$q_{i,m}(\mathbf{s}\mid X_{\mathcal{O}_i})
=\mathcal{N}(\mathbf{s};\,\boldsymbol{\mu}_{i,m},\,
\mathrm{diag}(\boldsymbol{\sigma}^{2}_{i,m}))$.
Combined with a standard normal prior $p(\mathbf{s})=\mathcal{N}(\mathbf{0},\mathbf{I})$,
the product $p(\mathbf{s})\prod_{m=1}^{M}q_{i,m}$ yields a closed-form
Gaussian $q_i(\mathbf{s}\mid X_{\mathcal{O}_i})
=\mathcal{N}(\boldsymbol{\mu}_i,\,\mathrm{diag}(\boldsymbol{\sigma}_i^{2}))$
via precision aggregation:
\begin{equation}
\boldsymbol{\sigma}_i^{-2}=\mathbf{1}+\sum_{m=1}^{M}\boldsymbol{\sigma}_{i,m}^{-2},
\qquad
\boldsymbol{\mu}_i=\boldsymbol{\sigma}_i^{2}\odot\sum_{m=1}^{M}
\big(\boldsymbol{\mu}_{i,m}\odot\boldsymbol{\sigma}_{i,m}^{-2}\big).
\end{equation}
Note that all $M$ experts participate regardless of missingness; experts
backed by imputed features naturally exhibit larger
$\boldsymbol{\sigma}_{i,m}^{2}$ and thus contribute less precision to the
fused result. 

\noindent\textbf{Prediction via Monte Carlo (MC) Sampling.}
We draw $L$ samples from the fused posterior and average the classifier outputs:
\begin{equation}
\mathbf{p}_i=\frac{1}{L}\sum_{k=1}^{L}\mathrm{Softmax}(W\mathbf{s}_i^{(k)}+b),
\qquad \mathbf{s}_i^{(k)}\sim q_i(\mathbf{s}\mid X_{\mathcal{O}_i}).
\end{equation}  

\subsection{Loss Function}
We train the model with the cross-entropy loss. To regularize the fused distribution and avoid over-confident experts, we add a Kullback-Leibler (KL) Divergence penalty to the standard normal prior \(p(\mathbf{s})=\mathcal{N}(\mathbf{0},\mathbf{I})\):
\begin{equation}
\mathcal{L}
=\frac{1}{N}\sum_{i=1}^{N}\Big(
\mathcal{L}_{\mathrm{CE}}(\mathbf{p}_i,y_i)
+\beta\,D_{\mathrm{KL}}\!\big(q_i(\mathbf{s}\mid X_{\mathcal{O}_i}) \,\|\, p(\mathbf{s})\big)
\Big),
\end{equation}
where \(\mathbf{p}_i\) is the predictive probability (Sec.~\ref{sec:uapoe}) and \(q_i(\mathbf{s}\mid X_{\mathcal{O}_i})\) is the UA-PoE fused Gaussian, $\beta$ is a hyper-parameter.
\section{Experimental Results}

\subsection{Dataset and Implementation Details}
\noindent \textbf{Datasets.}  We evaluate on baseline ADNI~\cite{mueller2005alzheimer} and OASIS-3~\cite{lamontagne2019oasis}. ADNI contains T1-sMRI, FDG-PET, Amyloid-PET, and tabular variables (Age, gender, education, marital status, MMSE, APOE4, ADNI\_EF, ADNI\_MEM); OASIS-3 contains T1-sMRI, Flair-MRI, Amyloid-PET, and tabular variables (Age, gender, education, APOE4, MMSE), which are commonly used in AD multimodal learning~\cite{xue2024ai,qiu2022multimodal,yang2025adfound}. We follow~\cite{xue2024ai} for 3D image preprocessing and Min--Max normalize tabular features; all volumes are resized to \(128^3\). ADNI contains 825/980/429 Cognitively Normal (CN)/Mild Cognitive Impairment (MCI)/AD subjects (N=2,234), and OASIS-3 has 791/197 CN/AD subjects (N=988). We split complete subjects into dev/test (1:1), then split (dev + incomplete) into train/val (6:1) stratified by diagnosis, resulting in 1504/259/471 (ADNI) and 757/127/104 (OASIS-3) for train/val/test. Train sets exhibit realistic missingness (e.g., 37.4\% AV45 available), while test sets are complete to simulate arbitrary missing-modality subsets at inference (Fig.~\ref{fig:intro}).

\noindent\textbf{Implementation Details.}
We train with Adam (LR \(2\times10^{-4}\)), 15-epoch warm-up + cosine decay, batch size 32, up to 100 epochs with early stopping (patience 40); \(\beta=0.001\) unless specified. Experiments use two NVIDIA RTX 4090 GPUs.
\begin{table}[!h]
\centering
\begingroup

\setlength{\origtabcolsep}{\tabcolsep}
\addtolength{\tabcolsep}{0.2em}

\newcommand{\mstd}[2]{#1\,{\scalebox{0.8}{(#2)}}}

\caption{Performance comparison (mean (std), \%) on ADNI and OASIS-3 for average metrics on 15 modality combinations, with computational efficiency. The \textcolor{orange!75}{orange} background refers to the best one, and the second-best results are \underline{underlined}.}

\resizebox{\textwidth}{!}{%
\begin{tabular}{@{\hspace{\origtabcolsep}}c@{\hspace{2\origtabcolsep}}ccccccccc}
\toprule
\multirow{2}{*}{\raisebox{-2.3ex}{\textbf{Methods}}} &
\multicolumn{3}{c}{\textbf{CN vs. MCI vs. AD}} &
\multicolumn{3}{c}{\textbf{CN vs. AD}} &
\multicolumn{3}{c}{\textbf{Computational Efficiency}} \\
\cmidrule(lr){2-4}\cmidrule(lr){5-7}\cmidrule(lr){8-10}
\multicolumn{1}{c}{} &
\textbf{ACC} & \textbf{AUC} & \textbf{Macro-F1} &
\textbf{ACC} & \textbf{AUC} & \textbf{F1\textsubscript{AD}} &
\makecell[c]{\textbf{Params}\\\textbf{(M)}} &
\makecell[c]{\textbf{FLOPs}\\\textbf{(G)}} &
\makecell[c]{\textbf{Speed}\\\textbf{(ms)}} \\
\midrule
\makecell[c]{mmFormer \cite{zhang2022mmformer}}      & \mstd{51.7}{2.1} & \mstd{68.4}{1.1} & \mstd{44.6}{2.6} & \mstd{86.8}{2.1} & \mstd{72.8}{6.2} & \mstd{40.3}{12.2} & 26.4  & 65.2  & 31.7 \\
\makecell[c]{ShaSpec \cite{wang2023multi}}        & \mstd{56.9}{1.9} & \mstd{73.6}{1.4} & \mstd{54.1}{1.0} & \mstd{87.7}{0.0} & \mstd{88.2}{2.4} & \mstd{51.0}{13.5} & 144.1 & 482.4 & 43.9 \\
\makecell[c]{AnyMod \cite{feng2024unified}}        & \mstd{59.7}{1.1} & \mstd{76.6}{1.3} & \mstd{56.8}{3.6} & \mstd{85.5}{1.6} & \mstd{82.3}{4.8} & \mstd{51.5}{4.5}  & \cellcolor{orange!15}11.9 & 92.4  & 129.8 \\
\makecell[c]{M3Care~\cite{zhang2022m3care}}         & \mstd{59.7}{1.1} & \underline{\mstd{78.7}{1.5}} & \underline{\mstd{59.9}{2.0}} & \mstd{88.5}{1.7} & \underline{\mstd{88.7}{0.7}} & \mstd{52.9}{6.2}  & 19.2  & \cellcolor{orange!15}18.4  & 11.9 \\
\makecell[c]{FuseMoE \cite{han2025fusemoe}}         & \mstd{58.5}{1.2} & \mstd{77.6}{0.3} & \mstd{57.0}{2.0} & \mstd{88.9}{1.7} & \mstd{86.6}{1.3} & \mstd{57.7}{5.9}  & 118.8 & \underline{18.5}  & \cellcolor{orange!15}2.5 \\
\makecell[c]{Flex-MoE \cite{yun2025flex}}            & \mstd{57.1}{1.7} & \mstd{74.3}{1.4} & \mstd{48.1}{2.5} & \mstd{88.7}{2.3} & \mstd{87.5}{1.0} & \mstd{57.3}{7.0}  & \underline{13.8}  & \cellcolor{orange!15}18.4  & 28.0 \\
\makecell[c]{MoE-retriever \cite{yun2025generate}}  & \underline{\mstd{60.6}{3.1}} & \mstd{78.3}{1.9} & \mstd{57.5}{2.6} & \underline{\mstd{89.8}{0.3}} & \mstd{86.4}{1.9} & \underline{\mstd{58.0}{2.8}}  & 15.1  & \cellcolor{orange!15}18.4  & 74.2 \\
\makecell[c]{\textbf{PRA-PoE (Ours)}} &
\cellcolor{orange!15}\mstd{63.9}{0.4} &
\cellcolor{orange!15}\mstd{81.4}{0.3} &
\cellcolor{orange!15}\mstd{61.7}{0.8} &
\cellcolor{orange!15}\mstd{90.3}{0.6} &
\cellcolor{orange!15}\mstd{89.5}{0.5} &
\cellcolor{orange!15}\mstd{64.3}{0.9} &
\underline{13.8} &
\cellcolor{orange!15}18.4 &
\underline{8.1} \\
\bottomrule
\end{tabular}%
}

\label{tab:adni_oasis3_main_percent_1dp}
\endgroup
\end{table}

\begin{table}[!h]
\centering
\begingroup

\setlength{\origtabcolsep}{\tabcolsep}
\addtolength{\tabcolsep}{0.2em}

\newcommand{\mstd}[2]{#1\,{\scalebox{0.8}{(#2)}}}

\caption{ACC comparison under 8 clinical common modality combinations on ADNI dataset.}
\resizebox{\textwidth}{!}{%
\begin{tabular}{  @{\hspace{\origtabcolsep}}c@{\hspace{2\origtabcolsep}}
  c@{\hspace{2\origtabcolsep}}
  c@{\hspace{2\origtabcolsep}}
  c@{\hspace{\origtabcolsep}}|
  cccccccc}
\toprule
\multicolumn{4}{c|}{\textbf{Modalities}} & \multicolumn{8}{c}{\textbf{ACC (\%)}} \\
\midrule
Tab& T1 & FDG & Amyloid &
\makecell[c]{mmFormer} &
\makecell[c]{ShaSpec} &
\makecell[c]{AnyMod} &
\makecell[c]{M3Care} &
\makecell[c]{FuseMoE} &
\makecell[c]{Flex-MoE} &
\makecell[c]{MoE-retr} &
\makecell[c]{\textbf{PRA-PoE}} \\
\midrule

$\checkmark$ & & & &
\mstd{56.8}{5.0} &
\cellcolor{orange!15}\mstd{67.9}{3.5} &
\mstd{64.2}{2.4} &
\mstd{60.3}{2.9} &
\mstd{63.6}{2.5} &
\mstd{65.2}{3.3} &
\mstd{61.3}{7.6} &
\underline{\mstd{65.3}{0.5}} \\

$\checkmark$ & $\checkmark$ & & &
\mstd{56.3}{4.0} &
\mstd{56.3}{3.7} &
\underline{\mstd{64.8}{1.8}} &
\mstd{63.4}{5.4} &
\mstd{61.1}{4.6} &
\mstd{64.8}{2.3} &
\mstd{63.7}{7.5} &
\cellcolor{orange!15}\mstd{68.7}{1.6} \\

$\checkmark$ & & $\checkmark$ & &
\mstd{60.6}{2.0} &
\mstd{59.9}{1.7} &
\mstd{67.4}{0.5} &
\mstd{65.6}{1.1} &
\mstd{68.0}{0.9} &
\underline{\mstd{69.1}{0.7}} &
\mstd{65.0}{2.6} &
\cellcolor{orange!15}\mstd{70.1}{0.5} \\

$\checkmark$ & & & $\checkmark$ &
\mstd{53.3}{2.8} &
\mstd{58.5}{3.3} &
\underline{\mstd{65.9}{0.6}} &
\mstd{61.1}{3.3} &
\mstd{59.7}{3.1} &
\mstd{62.1}{2.3} &
\mstd{62.0}{6.2} &
\cellcolor{orange!15}\mstd{68.4}{0.5} \\

$\checkmark$ & $\checkmark$ & $\checkmark$ & &
\mstd{60.2}{1.8} &
\mstd{60.7}{1.9} &
\mstd{66.8}{1.0} &
\mstd{66.7}{2.7} &
\mstd{68.3}{1.2} &
\cellcolor{orange!15}\mstd{70.3}{0.4} &
\mstd{67.3}{2.1} &
\underline{\mstd{69.5}{1.2}} \\

$\checkmark$ & $\checkmark$ & & $\checkmark$ &
\mstd{53.8}{3.0} &
\mstd{57.7}{4.6} &
\underline{\mstd{66.2}{0.6}} &
\mstd{64.1}{4.4} &
\mstd{60.4}{5.4} &
\mstd{66.0}{3.4} &
\mstd{64.0}{5.0} &
\cellcolor{orange!15}\mstd{69.1}{0.8} \\

$\checkmark$ & & $\checkmark$ & $\checkmark$ &
\mstd{59.9}{2.1} &
\mstd{61.7}{3.3} &
\mstd{67.3}{0.8} &
\mstd{65.7}{1.3} &
\mstd{69.4}{1.5} &
\underline{\mstd{69.8}{0.8}} &
\mstd{69.3}{2.1} &
\cellcolor{orange!15}\mstd{71.9}{2.1} \\

$\checkmark$ & $\checkmark$ & $\checkmark$ & $\checkmark$ &
\mstd{61.0}{1.9} &
\mstd{60.7}{3.6} &
\mstd{66.9}{1.4} &
\mstd{65.8}{2.8} &
\mstd{68.8}{1.8}&
\underline{\mstd{69.5}{0.1}} &
\mstd{69.0}{4.0} &
\cellcolor{orange!15}\mstd{72.1}{0.8} \\

\midrule
\multicolumn{4}{c|}{Average} &
\mstd{57.7}{2.7} & \mstd{60.4}{2.5} & \mstd{66.2}{1.0} & \mstd{64.1}{0.7} &
\mstd{64.9}{1.2} & \mstd{67.1}{1.2} & \underline{\mstd{65.2}{4.1}} &
\cellcolor{orange!15}\mstd{69.4}{1.3} \\

\bottomrule
\end{tabular}%
}
\label{tab:acc_adni}
\endgroup
\end{table}
\subsection{Comparison with State-of-the-art Methods.}
We evaluate all methods under arbitrary missing-modality settings for CN vs. MCI vs. AD (ADNI) and CN vs. AD (OASIS-3). For each test subject, we randomly drop modalities to form different modality subsets and report the average performance over all 15 non-empty modality combinations. We compare PRA-PoE with state-of-the-art approaches for incomplete multimodal learning, including mmFormer~\cite{zhang2022mmformer} and ShaSpec~\cite{wang2023multi} (adapted to classification by replacing decoders with classification heads), M3Care~\cite{zhang2022m3care}, AnyMod~\cite{feng2024unified}, and recent flexible Mixture-of-Experts (MoE) frameworks (FuseMoE~\cite{han2025fusemoe}, Flex-MoE~\cite{yun2025flex}, and MoE-retriever~\cite{yun2025generate}). All experiments are repeated three times with different random seeds, and results are reported as mean(standard deviation).

The overall quantitative results are summarized in Table~\ref{tab:adni_oasis3_main_percent_1dp}. For CN vs. MCI vs. AD, PRA-PoE achieves the best average performance across modality combinations, with an average ACC of 63.9\%, AUC of 81.4\%, and Macro-F1 of 61.7\%, outperforming MoE-retriever (the best baseline in ACC) by relative improvements of \textbf{5.4\%} in ACC, \textbf{4.0\%} in AUC, and \textbf{7.8\%} in Macro-F1 score on the ADNI dataset. For CN vs. AD,  PRA-PoE outperforms other methods with an average ACC of 90.3\%, AUC of 89.5\%, and F1\textsubscript{AD} of 64.3\% on the OASIS-3 dataset. Due to page limitations, we only present the Accuracy of 8 clinically common modality combinations in ADNI and OASIS-3 in Table~\ref{tab:acc_adni} and Table~\ref{tab:oasis3_acc}. In addition, Table~\ref{tab:adni_oasis3_main_percent_1dp} reports model complexity: PRA-PoE is parameter-efficient (13.8M) and computationally lightweight (18.4G FLOPs), while maintaining fast inference (8.1 ms/sample), making it suitable for practical low-latency deployment.
\begin{table}[!h]
\centering
\begingroup

\setlength{\origtabcolsep}{\tabcolsep}
\addtolength{\tabcolsep}{0.2em}

\newcommand{\mstd}[2]{#1\,{\scalebox{0.8}{(#2)}}}

\caption{ACC (\%) comparison under 8 clinical common modality combinations on OASIS-3 dataset.}
\resizebox{\textwidth}{!}{%
\begin{tabular}{  @{\hspace{\origtabcolsep}}c@{\hspace{2\origtabcolsep}}
  c@{\hspace{2\origtabcolsep}}
  c@{\hspace{2\origtabcolsep}}
  c@{\hspace{\origtabcolsep}}|
  cccccccc}
\toprule
\multicolumn{4}{c|}{\textbf{Modalities}} & \multicolumn{8}{c}{\textbf{ACC (\%)}} \\
\midrule
Tab & T1& Flair & Amyloid &
\makecell[c]{mmFormer} &
\makecell[c]{ShaSpec} &
\makecell[c]{AnyMod} &
\makecell[c]{M3Care} &
\makecell[c]{FuseMoE} &
\makecell[c]{Flex-MoE} &
\makecell[c]{MoE-retr} &
\makecell[c]{\textbf{PRA-PoE}} \\
\midrule

$\checkmark$ &  &  &  &
\mstd{85.6}{1.7} & \mstd{86.5}{6.7} & \cellcolor{orange!15}\mstd{90.4}{1.0} & \underline{\mstd{89.7}{2.8}} & \mstd{89.4}{1.0} & \mstd{88.5}{1.0} & \mstd{89.1}{2.1} & \cellcolor{orange!15}\mstd{90.4}{1.9} \\

$\checkmark$ & $\checkmark$ &  &  &
\mstd{85.6}{3.5} & \mstd{89.4}{1.0} & \mstd{86.2}{1.5} & \mstd{87.5}{3.3} & \mstd{88.8}{1.5} & \mstd{85.3}{4.5} & \underline{\mstd{90.7}{2.4}} & \cellcolor{orange!15}\mstd{91.7}{1.1} \\

$\checkmark$ &  & $\checkmark$ &  &
\mstd{82.7}{2.5} & \mstd{84.3}{2.4} & \mstd{84.6}{2.9} & \mstd{86.2}{2.4} & \mstd{85.3}{4.5} & \underline{\mstd{88.1}{2.8}} & \mstd{86.5}{1.9} & \cellcolor{orange!15}\mstd{89.7}{0.6} \\

$\checkmark$ &  &  & $\checkmark$ &
\mstd{87.5}{4.4} & \mstd{90.4}{1.0} & \mstd{90.4}{3.3} & \mstd{89.1}{3.1} & \cellcolor{orange!15}\mstd{92.9}{0.6} & \mstd{91.3}{1.7} & \mstd{90.4}{1.7} & \underline{\mstd{92.9}{2.2}} \\

$\checkmark$ & $\checkmark$ & $\checkmark$ &  &
\mstd{86.5}{1.0} & \mstd{87.8}{1.5} & \mstd{84.3}{1.5} & \mstd{87.8}{2.0} & \mstd{87.8}{2.9} & \mstd{86.2}{3.6} & \cellcolor{orange!15}\mstd{92.3}{1.7} & \underline{\mstd{91.3}{1.0}} \\

$\checkmark$ & $\checkmark$ &  & $\checkmark$ &
\mstd{88.8}{3.1} & \mstd{90.7}{2.0} & \mstd{89.7}{3.6} & \mstd{90.4}{2.5} & \cellcolor{orange!15}\mstd{92.6}{1.5} & \mstd{89.7}{1.5} & \underline{\mstd{92.3}{1.0}} & \mstd{92.0}{0.6} \\

$\checkmark$ &  & $\checkmark$ & $\checkmark$ &
\mstd{88.5}{1.0} & \mstd{87.5}{1.9} & \mstd{87.5}{5.4} & \mstd{90.7}{1.1} & \mstd{90.7}{2.4} & \mstd{92.0}{2.2} & \underline{\mstd{92.3}{0.0}} & \cellcolor{orange!15}\mstd{92.9}{1.1} \\

$\checkmark$ & $\checkmark$ & $\checkmark$ & $\checkmark$ &
\mstd{88.5}{1.7} & \mstd{88.5}{1.0} & \mstd{88.1}{4.9} & \mstd{90.1}{2.0} & \mstd{90.1}{2.4} & \mstd{89.7}{3.6} & \underline{\mstd{93.3}{1.0}} & \cellcolor{orange!15}\mstd{93.6}{0.6} \\

\midrule
\multicolumn{4}{c|}{Average} &
\mstd{86.7}{2.2} & \mstd{88.1}{0.3} & \mstd{87.7}{2.7} & \mstd{88.9}{1.7} &
\mstd{89.7}{1.4} & \mstd{88.9}{2.1} & \underline{\mstd{90.9}{0.8}} &
\cellcolor{orange!15}\mstd{91.8}{0.8} \\
\bottomrule
\end{tabular}%
}
\label{tab:oasis3_acc}
\endgroup
\end{table}

\begin{figure}[!h]
\centerline{\includegraphics[width=1\columnwidth]{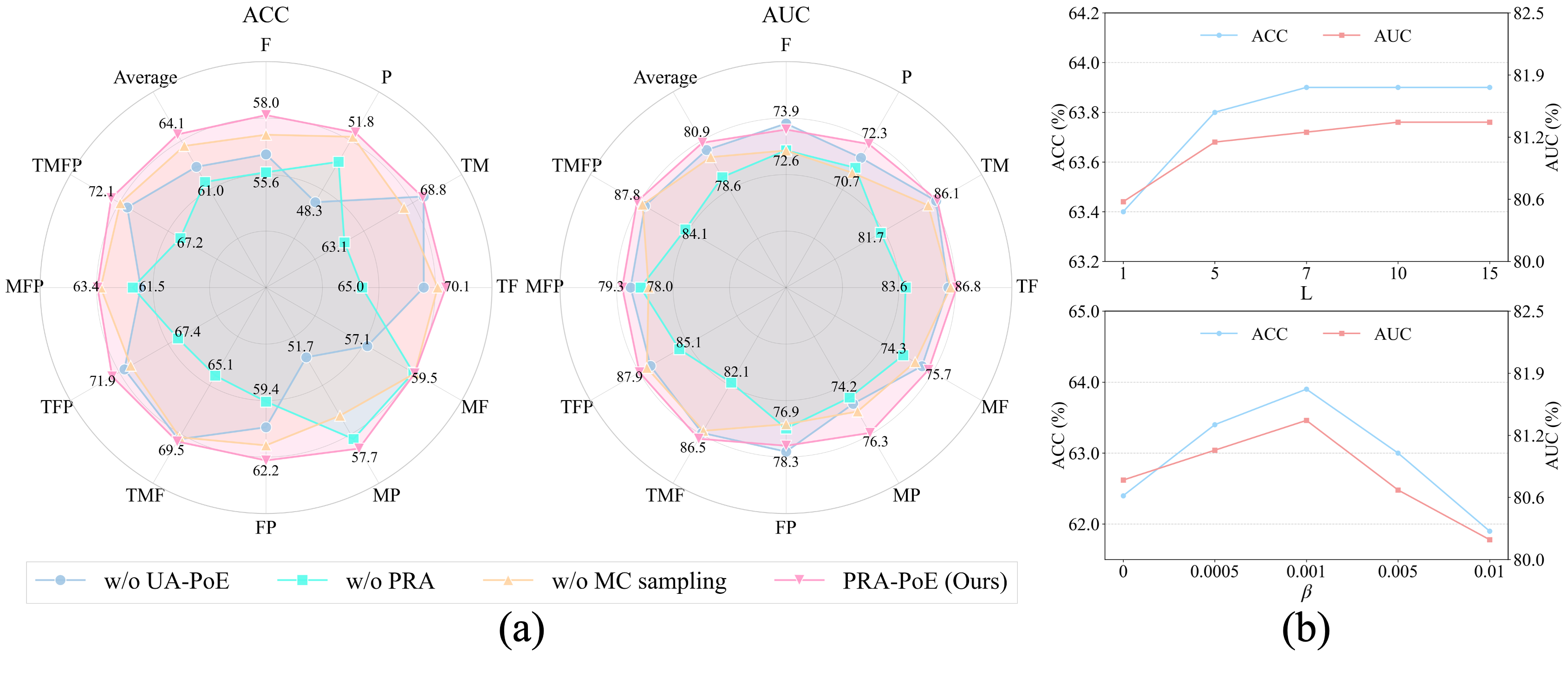}}
\caption{Ablation study (a) and hyperparameter sensitivity analysis (b).
} 
\label{fig:ablation}
\end{figure}

\subsection{Ablation Study and Sensitivity Analysis}
Fig.~\ref{fig:ablation} reports ablations and hyperparameter sensitivity on ADNI. In Fig.~\ref{fig:ablation}(a), PRA-PoE achieves the best average performance compared to variants without PRA, UA-PoE, or MC sampling across missingness patterns, showing all components contribute to robustness. Removing PRA yields the largest drop (Average ACC \(64.1\%\!\rightarrow\!61.0\%\)), indicating its key role in mitigating distribution shift, while removing UA-PoE also degrades performance (Average ACC \(64.1\%\rightarrow\!62.2\%\)), supporting uncertainty-aware fusion. In Fig.~\ref{fig:ablation}(b), increasing MC draws \(L\) improves performance and saturates at \(L\approx10\). The best KL weight is \(\beta=0.001\); larger \(\beta\) hurts performance, suggesting overly strong regularization limits discriminative representations.

\section{Conclusion}

In this paper, we propose PRA-PoE, a robust framework for AD classification under arbitrary modality missingness. By combining Prototype-anchored Representation Alignment to mitigate missingness-induced representation shift with uncertainty-aware precision fusion to attenuate unreliable modality estimates, PRA-PoE delivers robust predictions across diverse missing patterns. Experiments on two datasets demonstrate improved performance over strong baselines, indicating its potential for clinical application. Future work will extend the framework to longitudinal modeling and external multi-center validation to further assess generalizability.

\medskip\noindent\textbf{Acknowledgments.} 
\small 
This work was partially supported by RGC Collaborative Research Fund (No. C5055-24G), the Start-up Fund of The Hong Kong Polytechnic University (No. P0045999), the Seed Fund of the Research Institute for Smart Ageing (No. P0050946), and Tsinghua-PolyU Joint Research Initiative Fund (No. P0056509), and PolyU UGC funding (No. P0053716).

\medskip\noindent\textbf{Disclosure of Interests.} 
\small 
The authors have no competing interests.

%
%
%
%
\bibliographystyle{splncs04}
\bibliography{main}

\end{document}